\documentclass[10pt]{article}

\usepackage[a4paper,margin=1in]{geometry}
\usepackage{amsmath,amssymb}
\usepackage{graphicx}
\usepackage{booktabs}
\usepackage{hyperref}
\usepackage{algorithm}
\usepackage{algorithmic}
\usepackage{multirow}
\usepackage{xcolor}
\usepackage[numbers]{natbib}
\usepackage{caption}
\usepackage{subcaption}
\usepackage{authblk}
\usepackage{orcidlink}
\usepackage{wrapfig}
\usepackage{adjustbox}

\usepackage{tikz}
\usetikzlibrary{patterns, calc, decorations.pathreplacing}

\newtheorem{definition}{Definition}

\newcommand{\A}{\mathcal{A}}
\newcommand{\W}{\mathcal{W}}

\newcommand{\Tper}{T_{\text{per}}}

\graphicspath{{figures/}}

\hypersetup{
    colorlinks,
    linkcolor={red!50!black},
    citecolor={blue!50!black},
    urlcolor={blue!80!black}
}

\title{
LVSA: Training-Free Sparse Attention for Long Video Diffusion
}

\author[1]{Gael Glorian\,\orcidlink{0000-0002-0843-5987}\thanks{Corresponding author: \texttt{gael.glorian@huawei.com}}}
\author[1]{Ioannis Lamprou\,\orcidlink{0000-0001-5337-7336}}
\author[1]{Zhen Zhang\,\orcidlink{0009-0000-1130-8527}}
\author[1]{Yujie Yuan}
\author[2]{Hongsheng Liu\,\orcidlink{0000-0003-0509-7967}}

\affil[1]{\normalsize Distributed Parallel Technology Laboratory, Paris Research Center, Huawei Technologies France}
\affil[2]{\normalsize AI Framework and Data Technology Lab, Huawei Technologies Co., Ltd.}

\date{}

\begin{document}
\maketitle

\begin{abstract}
Dense self-attention is the compute and quality bottleneck of long-video diffusion inference: cost grows quadratically with the sequence length, and beyond the training horizon the model converges to near-static output, that is, ``frozen'' repetitive video. State of the art approaches are either too costly, e.g., they require retraining, or fail to satisfy both performance and quality objectives in a scalable manner. To this end, we introduce Long Video Sparse Attention (LVSA), a training-free model-agnostic block-sparse attention for video diffusion transformers that combines a structured window pattern with rotating global anchors, thus removing the fixed-grid bias which causes long-range temporal artifacts.
LVSA, combined with a FlashInfer kernel, reduces compute up to $3.17\times$ on Wan~2.1~1.3B at a $6\times$ horizon, $2.98\times$ on Wan~2.1~14B at a $6\times$ horizon, and $3.33\times$ on HunyuanVideo~1.5 at a $1.5\times$ horizon, compared to dense attention.
Beyond reducing compute, LVSA enables HunyuanVideo~1.5 generation at a $2\times$  horizon, which is otherwise out-of-memory on a single GPU. 
Moreover, LVSA provides speedups up to $2.41\times$ compared to RIFLEx \cite{zhao2025riflex} and $3.27\times$ compared to UltraViCo \cite{chen2025ultravico} on Wan~2.1~1.3B. 
To demonstrate applicability across diverse platforms, we apply LVSA on NPUs and achieve speedups up to $2.71\times$ on Wan~2.2~A14B and $3.24\times$ on Wan~2.1~1.3B compared to dense attention.
To evaluate quality in a fair way, we introduce VQeval, a tool properly scoring loopy video failures, which instead are rewarded in state of the art evaluators like VBench-Long \cite{yu2024vbench}. LVSA is quality-neutral for generation at training horizon length and quality-positive at extended lengths. 
\textbf{code}: \url{https://github.com/JiusiServe/LongVideoSparseAttention}
\end{abstract}

\section{Introduction}
\label{sec:intro}

Video diffusion transformers (DiTs) like Wan~\citep{wang2025wan} and HunyuanVideo~\citep{kong2024hunyuanvideo} have set new bars for text-to-video generation quality, but inference costs rise steeply with the number of generated frames as standard self-attention brings about quadratic compute. At the $14$-billion parameter scale, KV memory is pushed near the 80\,GB GPU envelope thus making longer video generation infeasible. Moreover, with respect to video quality, beyond the training horizon of $81$ frames for Wan and $129$ frames for HunyuanVideo, dense attention produces frozen or looping video, which is of very low quality by an observant's standards.

The above compute and quality challenges have captured the interest of the research community \cite{waseem2025video}. Sparse VideoGen~\citep{xi2025sparsevideogen, yang2025sparsevideogen2}, AdaSpa~\citep{xia2025adaspa}, Sliding Tile Attention~\citep{zhang2025sta}, and Radial Attention~\citep{li2025radial} all target the quadratic cost of video self-attention with training-free block or windowed patterns. Yet, the long-range temporal-repetition failures are still hard to eliminate. On the other hand, approaches on video extrapolation with quality preservation fail to deescalate the compute cost: RIFLEx~\citep{zhao2025riflex} modifies a single temporal RoPE frequency to extend the training horizon, while UltraViCo~\citep{chen2025ultravico} applies a per-pair logit decay via a fused Sage Attention \cite{zhang2025sageattentionaccurate8bitattention} kernel.
Note, in this paper, we consider a single-scene scenario.

In this work, we seek to address these shortcomings and harness the tradeoff between compute and quality. To do so, we make the following contributions:

\begin{itemize}
    \item We introduce Long-Video Sparse Attention (\textbf{LVSA}), a training-free block-sparse model-agnostic attention algorithm comprising novel rotating sparse patterns and expanded adaptive window logic.
    \item We introduce a custom evaluation benchmark, namely \textbf{VQeval}, to properly score loopy video failures, in contrast to state of the art VBench-Long \cite{yu2024vbench}.
    \item We experimentally validate the efficacy of our approach across three architecturally distinct video DiTs for inference on a single 80GB GPU. LVSA combined with a FlashInfer kernel delivers a $3.17\times$ speedup on Wan~2.1~1.3B, $2.98\times$ on Wan~2.1~14B, and $3.33\times$ on HunyuanVideo~1.5 at the longest tested generation horizon per model, while significantly outscoring dense attention on VQeval at a $6\times$ horizon on both Wan models. LVSA additionally enables HunyuanVideo~1.5 generation at a $2\times$ horizon (257 frames), where dense attention is infeasible due to memory exhaustion. LVSA outperforms UltraViCo and RIFLEx both in compute (up to $3.27\times$) and quality.
    \item We showcase the efficacy of LVSA for video generation on NPUs. We achieve a $2.71\times$ speedup on Wan~2.2~A14B and $3.24\times$ speedup on Wan~2.1~1.3B at a $6\times$ horizon, with good video quality.
    \item We include our implementation as a plugin in a popular open-source platform. 
\end{itemize}

\section{Method}
\label{sec:method}

A video diffusion transformer operates on a latent video tensor patchified into a sequence of $N = T \cdot P$ tokens, each of dimension $d$, where $T$ is the number of latent temporal frames, from now on simply referred to as \textit{frames}, and $P = H_p \cdot W_p$ is the number of spatial patches (height times width) per frame. Below, let $t \in \{0, 1, \ldots, T-1\}$ denote the $t$-th frame and $q_{t,p}, k_{t,p}, v_{t,p}$ denote the $p$-th query, key, and value, tokens for frame $t$, where $p \in \{0,1, \ldots, P-1\}$. 
The (spatio-temporal) self-attention formula for a query token $q_{t,i}$ is

\begin{equation}
\label{eq:spatio-temporal-attention}
\text{Attn}(q_{t,i}) = \sum_{\tau=0}^{T-1}\sum_{p=0}^{P-1} \frac{\exp(q_{t,i} \cdot k_{\tau,p} / \sqrt{d})}{\sum_{\tau'=0}^{T-1}\sum_{p'=0}^{P-1} \exp(q_{t,i} \cdot k_{\tau',p'} / \sqrt{d})} \cdot v_{\tau,p}.
\end{equation}

The (dense) attention cost for $\text{Attn}(q_{t,i})$ is $O(N d) = O(TP d)$. The formula must be computed for each $i$ and $t$, which leads to a prohibitive $O(N^2d)$ complexity for long video generation. Note that each query frame attends to all other frames in the temporal dimension $T$. Let us formalize and generalize this notion of per-frame attention.

\begin{definition}\label{def:attend}
 For a query frame $t$, the set of frames it attends to is denoted by $\A(t) \subseteq \{0, 1, \ldots, T-1\}$. In dense attention, $\A(t) = \{0, 1, \ldots, T-1\}$, that is, $t$ attends to all frames.
\end{definition}

To enable long video generation of high quality quickly, we introduce sparsity logic. We seek to restrict the set of frames a query frame attends to. Thus, we perform fewer computations, yet in a smart way to avoid sacrificing quality.  We generalize the above self-attention formula to: 
\begin{equation}
\label{eq:spatio-temporal-attention-sparse}
\text{Attn}(q_{t,i}) = \sum_{\tau \in \A(t)}\sum_{p=0}^{P-1} \frac{\exp(q_{t,i} \cdot k_{\tau,p} / \sqrt{d})}{\sum_{\tau'\in \A(t)}\sum_{p'=0}^{P-1} \exp(q_{t,i} \cdot k_{\tau',p'} / \sqrt{d})} \cdot v_{\tau,p}.
\end{equation}

Note the complexity to compute $\text{Attn}(q_{t,i})$ is $O(|\A(t)| P d)$.
The question that now remains is to define $\A(t)$ for each frame $t$.

Let us now define our attention pattern for each query frame $t$, which comprises two components. 
To maintain quality throughout time, we let each query frame attend to a set of (global) frames at key times during the whole time horizon. Also, each frame attends to a small (local) window of frames surrounding it temporally. Overall, we wish for each frame to attend to a sensible yet small number of frames, both locally and globally, such that compute is reduced, while quality is not compromised.

\begin{definition}\label{def:lvsa-attend}
The Long Video Sparse Attention (LVSA) formula is the spatio-temporal attention formula defined in Equation~\ref{eq:spatio-temporal-attention-sparse} with $\A(t) = G \,\cup\, \W(t)$, where
\begin{itemize}
    \item $G = \{t \;|\; t = 0, 1, \ldots, T-1 \;\land\; t \bmod \Tper = 0\}$ are equidistant global frames at a period $\Tper \in \mathbb{N}$,
    \item $\W(t) = \{t' \;|\; w_{\text{lo}}(t) \leq t' \leq w_{\text{hi}}(t)\}$ is a local window of frames around $t$ with $w_{lo}(t) = \max\{0, t - W\}$ and $w_{hi}(t) = \min\{T-1, t+W\}$, where $W \in \mathbb{N}$ is the window size and we assume $2W + 1 \le T$.
\end{itemize}
\end{definition}

\begin{figure}[ht]
\centering
\begin{minipage}[b]{0.495\columnwidth}
  \centering
  \includegraphics[width=\linewidth]{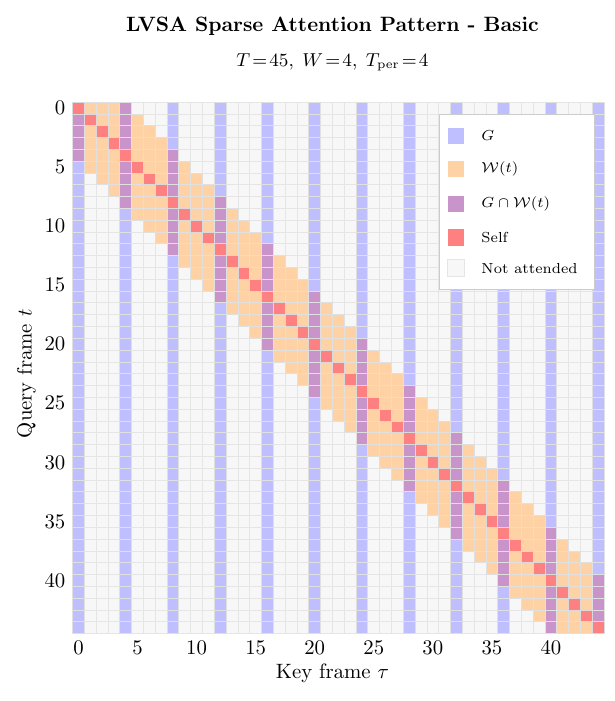}
  \subcaption{Basic window}
  \label{fig:pattern_base}
\end{minipage}\hfill
\begin{minipage}[b]{0.495\columnwidth}
  \centering
  \includegraphics[width=\linewidth]{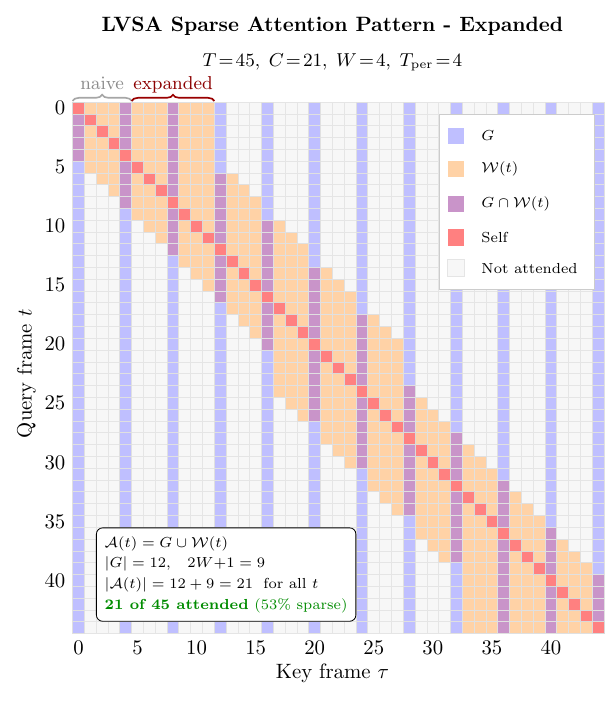}
  \subcaption{Expanded window}
  \label{fig:pattern_expanded}
\end{minipage}
\caption{Basic versus expanded window pattern. The basic adaptive window (a) wastes attention budget when the local window overlaps global frames, leaving the per-query attended set below the target $C$. Expanded bounds (b) account for this overlap by extending the window when needed, so every query frame attends to $|\mathcal{A}(f)| = |G| + \min(2W+1, T-|G|) \approx C$ unique frames.}
\label{fig:patterns}
\end{figure}

For a visual depiction of Definition~\ref{def:lvsa-attend}, see Figure~\ref{fig:pattern_base}. Note frame $0$ is always a global anchor, thus ensuring queries always attend to scene-establishing content.

Following the above definition, the user must specify $\Tper$ and $W$ to compute the formula on the globally and locally attended frames. The question arises on how to select these parameters. For simplicity of exposition we present LVSA with a single global set $G$ of equidistant frames; an extension including more dedicated initial anchors is a straightforward generalization.

To avoid fluctuations in compute, and maintain a constant budget per query frame, we fix a target budget $C$ and let $|\A(t)| \approx C$, for each frame $t$, with deviation bounded by integer keyframe-spacing rounding (at most $\pm 2$ frames across the configurations we evaluate). By default, we let $C$ be equal to the reference frame count of the trained model, e.g., $C = \frac{81-1}{4} +1 = 21$ frames for Wan~2.1~1.3B, as Wan is trained on an $81$-frame horizon and has its variational auto encoder (VAE) factor set to $4$. Intuitively, since the model is trained on a budget of $C$ frames, we use the same budget for inference. Using a smaller budget would improve efficiency, yet lower quality, while using a larger budget would be costly and eventually intractable. Overall, the complexity of $\text{Attn}(q_{t,i})$ becomes $O(CPd)$, for each frame $t$ and patch $i$, yielding a total complexity of $O(TCP^2d)$. The latter is asymptotically \textit{linear} in the number of frames $T$, thus enabling long video generation.

One could allocate the attention budget $C$ in any way, as long as $|\mathcal{A}(t)| \approx C$ for all $t$. In our case, we assume $W$ is already fine-tuned, see Section~\ref{sec:exp}. We allocate the remaining budget to the periodic global frames in $G$ and respectively set $\Tper = \left\lceil\frac{T}{C - (2W+1)}\right\rceil$. This assignment applies to the practical case where $C > 2W+1$. Since $\Tper$ is an integer, the realized $|G| = \lceil T / \Tper \rceil$ may differ from the target $C - (2W+1)$ by up to $\Tper - 1$ frames; in our experiments this is at most two frames in either direction.

\paragraph{Overlapping frames.}
\label{sec:method:expanded}

We now consider the case where $G \cap \mathcal{W}(t) \neq \emptyset$ and so some frames are included in both sets.
A naive local window, as given in Definition~\ref{def:lvsa-attend}, clips at sequence boundaries, giving edge frames a smaller attention budget than interior ones. A simple adaptive window shifts the range appropriately in order to maintain a constant window size for edge cases by setting $w_{\text{lo}}'(t) = \max(0, \min(t - W, T - 1 - 2W))$ and $w_{\text{hi}}'(t) = \min(T - 1, \max(t + W, 2W))$. Thus, every frame attends to exactly $2W + 1$ window frames. 

However, when window frames overlap global frames, the effective number of unique non-global frames in the window is reduced, wasting attention budget.
We introduce \textit{expanded window bounds} to compensate for this overlap, see Algorithm~\ref{alg:expanded} and Figure~\ref{fig:pattern_expanded}.

\begin{algorithm}[ht]
\caption{Expanded window bounds for $\mathcal{W}(t)$}
\label{alg:expanded}
\begin{algorithmic}[1]
\REQUIRE Frame $t$, global frames $G$, window size $W$, total frames $T$
\ENSURE Assign expanded windows bounds $(w_{\text{lo}}, w_{\text{hi}})$
\STATE $(l, h) \gets (w_{\text{lo}}'(t), w_{\text{hi}}'(t))$
\STATE \textit{target} $\gets \min(2W{+}1,\ T - |G|)$
\STATE \textit{nonglobal} $\gets |\{t' \in [l, h] \,|\, t' \notin G\}|$
\WHILE{\textit{nonglobal} $<$ \textit{target}$\;\land\; (l > 0 \;\lor\; h < T{-}1)$}
  \STATE extend the side with the most room by 1; increment \emph{nonglobal} if the new frame is $\notin G$
\ENDWHILE
\STATE $(w_{\text{lo}}, w_{\text{hi}}) \gets (l,h)$
\end{algorithmic}
\end{algorithm}

The cost of Algorithm~\ref{alg:expanded} is negligible, since in practice only a few iterations of the while loop will take place. The algorithm runs on CPU during metadata building and does not enter the attention critical path. With expanded bounds, the total unique attended frames satisfy
$|\mathcal{A}(t)| = |G| + \min(2W + 1, T - |G|)$
for each frame $t$, giving a uniform per-frame attention budget across the sequence.
For any $T \geq 2W + 1$, every query frame $t$ has $|\mathcal{A}(t)| \in [C - \delta, C + \delta]$ with $\delta \leq \Tper - 1$, and $|\mathcal{A}(t)|$ is constant across $t$ for fixed $G$.
Empirically, across all configurations used in our experiments
, the loop runs at most $8$ iterations per call (mean $1.21$) and the full per-frame call takes $\approx 1.4\,\mu s$ on a single CPU core. A complete metadata rebuild for a denoising step takes less than $200\,\mu s$, which is negligible compared to the GPU attention kernel.

\paragraph{Rotating periodic global frames.}
\label{sec:method:rotating}

Fixing the periodic global frames in $G$ allows for maintaining information throughout the video duration. Nonetheless, it also creates a persistent bias: these frames are always attended to globally, while intermediate frames are only observed through local windows. Over the course of $S$ denoising steps, the model's representation of frames not in $G$ is systematically impoverished. At extended lengths, this manifests as long-range temporal artifacts, e.g., repetition, identity drift, etc.

To address this, we introduce \textit{rotating periodic global frames}. At denoising step $s=0,1,\ldots,S-1$, the global set $G^s$ is different than at previous steps. We shift the members of $G$ by $s$ modulo $\Tper$ positions and define the set as a function of the denoising step as

\begin{equation}
\label{eq:rotating_keyframes}
G^s = \big\{(s \bmod \Tper + i \cdot \Tper) \bmod T \;|\; i = 0, 1, \ldots, \lceil T / \Tper \rceil - 1 \big\}.
\end{equation}

\begin{wrapfigure}{r}{0.5\textwidth}
\centering
\includegraphics[scale=0.725]{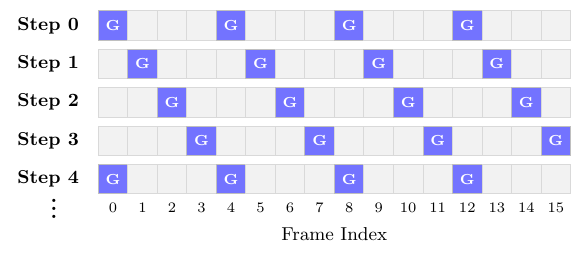}
\caption{Rotating periodic global frames with $T_{\text{per}} = 4$. The set $G^s$ shifts by one position per denoising step and wraps modulo $T$. Over any $T_{\text{per}}$ consecutive steps, each frame appears as a global anchor exactly once.}
\label{fig:rotating_frames}
\end{wrapfigure}

See Figure~\ref{fig:rotating_frames} for a visual depiction.

Two properties make this rotation principled. First, over any $\Tper$ consecutive denoising steps, every frame appears in $G^s$ for at least one value of $s$, so every frame serves as a global anchor at least once per cycle, eliminating the fixed-grid bias (when $T$ is not a multiple of $\Tper$, the last wrap may revisit at most $\Tper - 1$ frames within a cycle; this is empirically negligible at the keyframe-spacings used in our experiments). Second, the modular wrapping ensures $|G^s| = \lceil T / \Tper \rceil$ is constant across $s$, so the per-step attention budget does not change.

At each step we recompute the derived index structures 
for the rotated pattern. This is pure CPU index arithmetic over $T$ elements ($T$ ranges from $21$ at a $1\times$ Wan horizon to $121$ at a $6\times$ horizon) and takes less than $1$\,ms per step, which is negligible compared to the GPU attention kernel.

\section{Experiments on GPU}
\label{sec:exp}

\paragraph{Models.} We experiment with three architecturally distinct video DiTs: \textit{(i)} \textbf{Wan 2.1 T2V-1.3B} -- single-stream, 1D RoPE, T5 encoder, 40 steps, \textit{(ii)} \textbf{Wan 2.1 T2V-14B} -- same architecture at 14B parameters, 40 steps, and \textit{(iii)} \textbf{HunyuanVideo~1.5 (480p)} -- dual-stream, 3D RoPE, Qwen2.5-VL encoder, 50 steps.

\paragraph{Hardware.} A single GPU with 80\,GB, PyTorch 2.8/CUDA 12.8.

\paragraph{Video lengths.} $480 \times 832$ resolution. Wan: 81 ($1\times$ horizon), 161 ($2\times$), 241 ($3\times$), 321 ($4\times$), 401 ($5\times$), 481 ($6\times$) video frames. HunyuanVideo: 65 ($0.5\times$), 129 ($1\times$), 193 ($1.5\times$) and 257 ($2\times$) video frames.

\paragraph{Prompts, seed, and scheduler.} We test five diverse long descriptive prompts (around 500 tokens) with seed 16, classifier-free-guidance scale $5.0$, and each model's default scheduler. All cells in Sections~\ref{sec:exp:efficiency} and~\ref{sec:exp:quality} report mean $\pm$ standard deviation over the 5-prompt set.

\paragraph{Quality Metrics.} We report quality results on both \textbf{VBench-Long}~\citep{yu2024vbench} (subject consistency, temporal flickering, motion smoothness, background consistency, imaging quality) and \textbf{VQeval} (dynamic quality, loop quality, text alignment), a custom benchmark we introduce. The two benchmarks are complementary: VBench rewards inter-frame similarity, which scores static or collapsed videos highly, while VQeval's dynamic and loop dimensions explicitly penalize these failure modes.

\subsection{Computational Efficiency}
\label{sec:exp:efficiency}

\paragraph{Cross-model scaling.} We measure wall time across three models (Wan~2.1~1.3B, Wan~2.1~14B, HunyuanVideo~1.5) and three backends (dense attention, LVSA via scaled-dot-product attention (SDPA), LVSA via FlashInfer block-sparse kernel) over five long descriptive prompts per cell at generation horizons, which are multiples of each model's training horizon ($2\times$--$6\times$), see Table~\ref{tab:3model_scaling_wall}. At the longest tested horizon per model, LVSA with FlashInfer (LVSA-FI) achieves a $\mathbf{3.17\times}$ speedup on Wan~2.1~1.3B at a $6\times$ horizon (481 frames; 51\,min $\rightarrow$ 16\,min), $\mathbf{2.98\times}$ on Wan~2.1~14B at $6\times$ ($238$\,min $\rightarrow$ $80$\,min), and $\mathbf{3.33\times}$ on HunyuanVideo~1.5 at $1.5\times$ ($80$\,min $\rightarrow$ $24$\,min). The speedup is monotone in horizon and architecture-independent: the same three-model pattern emerges in single-stream/1D-RoPE (Wan) and dual-stream/3D-RoPE (HunyuanVideo) DiTs alike, driven by the quadratic-in-$T$ cost of dense self-attention. At native horizon ($1\times$), LVSA backends are at parity with dense (within $\pm 5\%$ wall time), reflecting the proportionally larger text-encoder cost when video self-attention is short.

\begin{figure}[t]
\centering
\includegraphics[width=\columnwidth]{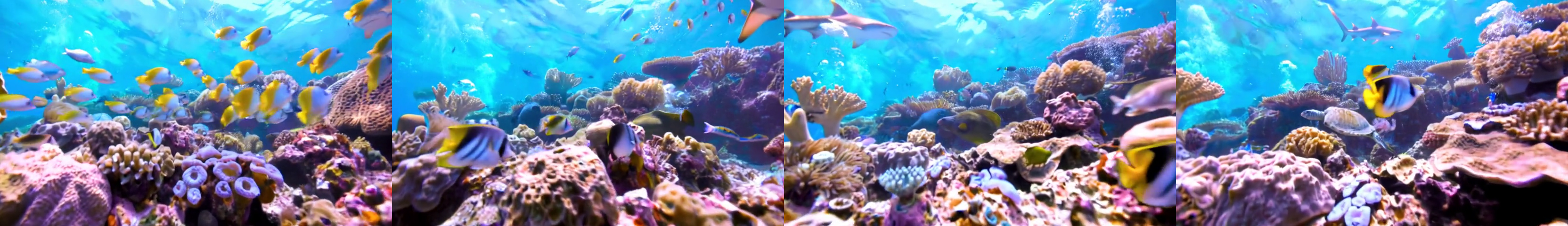}
\caption{HunyuanVideo~1.5 at $2\times$ horizon (257 frames), generated by LVSA-FI on a single 80GB GPU; dense attention is infeasible at this setting due to OOM (Table~\ref{tab:3model_scaling_wall}). Frames $32$, $96$, $160$, $224$ from the prompt \texttt{coral\_reef} (best-VQeval prompt at this cell, composite $62.9$).}
\label{fig:keyframes_hv15_2x}
\end{figure}

\paragraph{Feasibility at the GPU memory ceiling.} Beyond speedup, LVSA enables generation that is \emph{infeasible} with dense attention at a fixed GPU memory budget. On HunyuanVideo~1.5 at a $2\times$ horizon (257 frames), dense self-attention runs out of memory on a single 80GB GPU: the SDPA kernel attempts to allocate an additional $19.9$\,GB on top of a $74.0$\,GB resident process. LVSA at the same setting caps peak GPU memory at $60.3$\,GB (SDPA) / $60.4$\,GB (FlashInfer), leaving $\sim 19$\,GB of headroom and producing decoded video with VQeval composite $60.0$ / $58.5$ respectively --- numbers that have no dense counterpart at this hardware scale. Dense peak memory on HunyuanVideo~1.5 grows from $38.8$\,GB at a $0.5\times$ horizon to $67.4$\,GB at $1.5\times$ before exceeding the $80$\,GB budget at $2\times$. The asymmetric feasibility story --- HunyuanVideo~1.5 OOMs at $2\times$ extension while Wan~2.1~14B fits comfortably at $6\times$ (peak $57.8$\,GB on $481$ frames) --- is architectural, not parameter-count: both models are $\sim 14$B. HunyuanVideo~1.5 is dual-stream, with text-encoder tokens (Qwen2.5-VL + ByT5) participating in self-attention alongside video tokens, whereas Wan's text enters only via cross-attention. This extends HV's effective self-attention sequence by the encoder's output context per layer and pushes its attention activation matrix past the $80$\,GB budget at $2\times$, while Wan~2.1~14B's longer ($481$-frame) but text-free self-attention stays under. Figure~\ref{fig:keyframes_hv15_2x} shows representative frames from the HV~1.5 $2\times$ LVSA-FI output.

\begin{table}[t]
\centering
\footnotesize
\caption{Wall time in minutes per video generation on an 80GB GPU, mean over 5 long descriptive prompts. ``LVSA-FI'' = LVSA with FlashInfer kernel; speedup is LVSA-FI vs.\ dense attention. HunyuanVideo~(HV)~1.5 at a $2\times$ horizon is infeasible for dense attention, while LVSA fits in $\approx 60$\,GB.}
\label{tab:3model_scaling_wall}
\begin{tabular}{@{}llrrrrr@{}}
\toprule
\textbf{Model} & \textbf{Horizon} & \textbf{Frames} & \textbf{Dense} (min) & \textbf{LVSA} (min) & \textbf{LVSA-FI} (min) & \textbf{Speedup (LVSA-FI)} \\
\midrule
HV~1.5   & $0.5\times$           & 65  & $11.6$           & $5.9$  & $5.9$  & $1.97\times$ \\
HV~1.5   & $1\times$             & 129 & $37.5$           & $16.7$ & $16.4$ & $2.29\times$ \\
HV~1.5   & $1.5\times$           & 193 & $79.7$           & $26.4$ & $23.9$ & $3.33\times$ \\
HV~1.5   & $\mathbf{2\times}$    & 257 & \textbf{OOM}     & $57.9$ & $54.9$ & $\boldsymbol{\infty}$ \\
\midrule
Wan~2.1~1.3B & $1\times$             & 81  & $2.5$            & $3.1$  & $2.6$  & $0.96\times$ \\
Wan~2.1~1.3B & $2\times$             & 161 & $7.3$            & $6.5$  & $5.1$  & $1.42\times$ \\
Wan~2.1~1.3B & $4\times$             & 321 & $24.0$           & $13.1$ & $10.3$ & $2.32\times$ \\
Wan~2.1~1.3B & $\mathbf{6\times}$    & 481 & $50.8$           & $19.9$ & $16.0$ & $\mathbf{3.17\times}$ \\
\midrule
Wan~2.1~14B  & $1\times$             & 81  & $12.6$           & $13.5$ & $13.2$ & $0.96\times$ \\
Wan~2.1~14B  & $2\times$             & 161 & $35.8$           & $28.5$ & $26.0$ & $1.38\times$ \\
Wan~2.1~14B  & $4\times$             & 321 & $114.0$          & $57.8$ & $52.6$ & $2.17\times$ \\
Wan~2.1~14B  & $\mathbf{6\times}$    & 481 & $237.9$          & $87.2$ & $79.8$ & $\mathbf{2.98\times}$ \\
\bottomrule
\end{tabular}
\end{table}

\begin{figure}[t]
\centering
\includegraphics[width=\columnwidth]{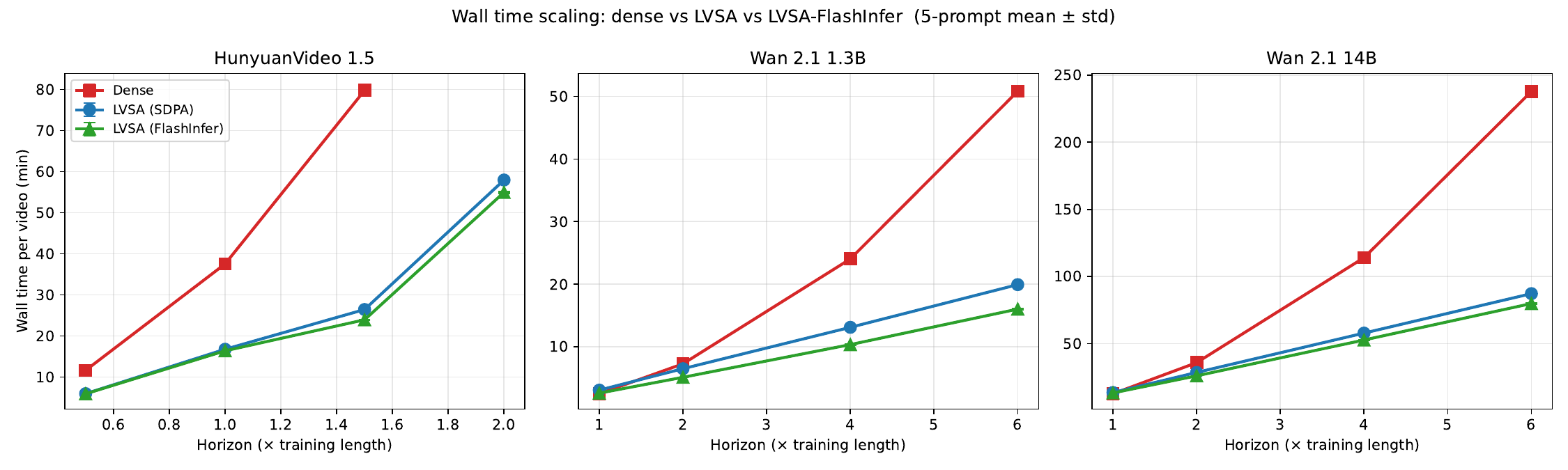}
\caption{Wall-time scaling across three video DiTs (Wan~2.1~1.3B, Wan~2.1~14B, HunyuanVideo~1.5) for dense attention, LVSA, and LVSA-FI. 
Speedup grows monotonically with the horizon for all three models; HunyuanVideo~1.5 at $2\times$ horizon (257 frames) has no dense point due to OOM on 80GB GPU.}
\label{fig:scaling_wall}
\end{figure}

\subsection{Video Quality}
\label{sec:exp:quality}

\paragraph{Quality at training horizon.} At each model's reference length ($1\times$), LVSA is quality-neutral with dense attention across all three architectures, see Table~\ref{tab:3model_quality}. VQeval composite is within $\pm 1.0$ of dense on all three models, and VBench-Long composite differences do not exceed $0.014$. Both LVSA backends (SDPA and FlashInfer) give equivalent quality at the training horizon, confirming that the attention pattern, and not the kernel choice, determines output quality.

\paragraph{Quality advantage at extended horizons.} Beyond the training horizon, LVSA's VQeval composite consistently outscores dense attention, with the gap widening monotonically with the horizon. On Wan~2.1~1.3B, the LVSA-FI advantage over dense grows from $+4.7$ at $2\times$ to $+11.6$ at $4\times$ and $+12.1$ at $6\times$. The same pattern holds on Wan~2.1~14B: $+3.8$ at $2\times$, $+9.7$ at $4\times$, $+12.2$ at $6\times$. Across the three architectures, dense attention's extrapolation failure beyond the training horizon means that dense converges to near-static output with reduced motion variation, which VQeval's dynamic and loop dimensions properly penalize. LVSA's sliding-window restriction acts as an implicit regularizer that preserves motion at extended horizons. The VBench-Long composite in Table~\ref{tab:3model_quality} tells the opposite story: dense scores rise at extended time horizons on both Wan models. The latter is a result of the static-rewarding bias of VBench's consistency dimensions discussed below.

\begin{table}[t]
\centering
\footnotesize
\caption{Quality metrics, mean over 5 long descriptive prompts. VQeval composite is on the $[0, 100]$ scale; VBench-Long composite is on $[0, 1]$. Bold marks the LVSA-FI cells where LVSA-FI matches or exceeds dense at $4\times$+ extension. HunyuanVideo~1.5 at $2\times$ has no dense baseline (OOM, Table~\ref{tab:3model_scaling_wall}). The two metrics diverge at Wan extension: dense's VBench composite \emph{rises} (rewarding its increasingly frozen output) while VQeval correctly tracks the lost motion.}
\label{tab:3model_quality}
\begin{tabular}{@{}llrrrrrrr@{}}
\toprule
 &  &  & \multicolumn{3}{c}{\textbf{VQeval composite}} & \multicolumn{3}{c}{\textbf{VBench-Long composite}} \\
\cmidrule(lr){4-6} \cmidrule(lr){7-9}
\textbf{Model} & \textbf{Horizon} & \textbf{Frames} & Dense & LVSA & LVSA-FI & Dense & LVSA & LVSA-FI \\
\midrule
HV~1.5   & $0.5\times$ & 65  & $56.9$ & $59.1$ & $57.2$ & $0.887$ & $0.884$ & $0.883$ \\
HV~1.5   & $1\times$   & 129 & $60.5$ & $61.2$ & $60.2$ & $0.893$ & $0.882$ & $0.879$ \\
HV~1.5   & $1.5\times$ & 193 & $61.0$ & $60.9$ & $62.5$ & $0.901$ & $0.897$ & $0.896$ \\
HV~1.5   & $2\times$   & 257 & OOM    & $60.0$ & $58.5$ & OOM     & $0.898$ & $0.896$ \\
\midrule
Wan~2.1~1.3B & $1\times$   & 81  & $56.5$ & $56.5$ & $56.7$ & $0.888$ & $0.888$ & $0.888$ \\
Wan~2.1~1.3B & $2\times$   & 161 & $55.3$ & $60.1$ & $60.0$ & $0.875$ & $0.869$ & $0.869$ \\
Wan~2.1~1.3B & $4\times$   & 321 & $50.9$ & $62.1$ & $\mathbf{62.5}$ & $0.885$ & $0.869$ & $0.869$ \\
Wan~2.1~1.3B & $6\times$   & 481 & $48.2$ & $60.1$ & $\mathbf{60.2}$ & $0.891$ & $0.852$ & $0.852$ \\
\midrule
Wan~2.1~14B  & $1\times$   & 81  & $57.2$ & $57.2$ & $58.2$ & $0.909$ & $0.908$ & $0.909$ \\
Wan~2.1~14B  & $2\times$   & 161 & $58.7$ & $62.5$ & $62.5$ & $0.904$ & $0.899$ & $0.900$ \\
Wan~2.1~14B  & $4\times$   & 321 & $55.0$ & $64.7$ & $\mathbf{64.8}$ & $0.899$ & $0.893$ & $0.892$ \\
Wan~2.1~14B  & $6\times$   & 481 & $52.7$ & $64.6$ & $\mathbf{64.9}$ & $0.886$ & $0.890$ & $\mathbf{0.890}$ \\
\bottomrule
\end{tabular}
\end{table}

\begin{figure}[t]
\centering
\begin{subfigure}{\columnwidth}
\centering
\includegraphics[width=\columnwidth]{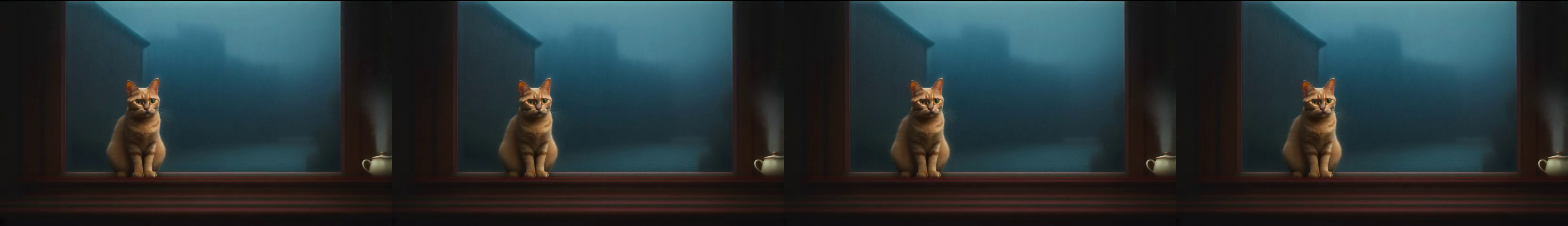}
\subcaption{Dense attention (VQeval $37.6$, \texttt{subject\_consistency} $0.991$).}
\end{subfigure}\\[4pt]
\begin{subfigure}{\columnwidth}
\centering
\includegraphics[width=\columnwidth]{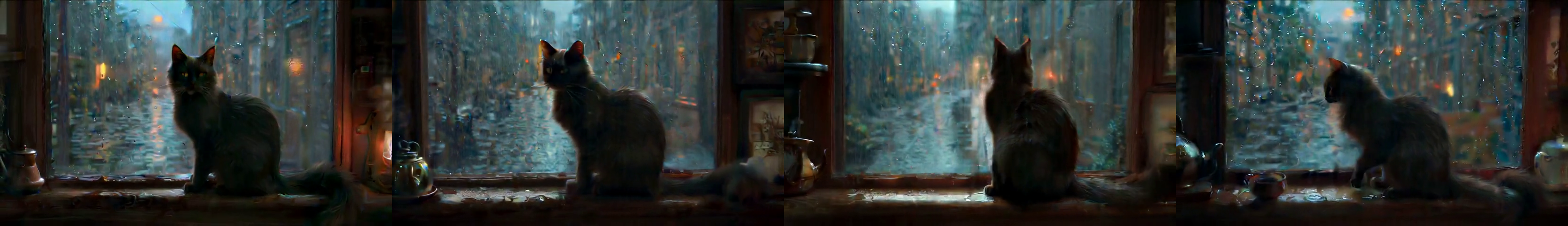}
\subcaption{LVSA (VQeval $53.1$, \texttt{subject\_consistency} $0.917$).}
\end{subfigure}
\caption{Wan~2.1~1.3B at a $6\times$ horizon (481 frames), prompt \texttt{cat\_window}, same seed. Frames $20$, $200$, $380$, $460$ shown for each backend. Dense converges to near-static output --- the cat barely moves across $\sim 440$ frames --- while LVSA produces genuine pose and lighting variation. This is the failure mode VBench-Long's \texttt{subject\_consistency} rewards and VQeval correctly penalizes.}
\label{fig:keyframes_wan13b_6x}
\end{figure}

\paragraph{VBench-Long behavior.} VBench-Long's composite increases for dense attention at extended horizons (Wan~2.1~1.3B $6\times$ dense at $0.891$ vs $4\times$ at $0.885$ vs $2\times$ at $0.875$), because two of its dimensions (\texttt{subject\_consistency}, \texttt{background\_consistency}) reward static video and dense attention's quality collapse at extended horizons produces increasingly frozen output, which VBench credits as ``consistent.'' At Wan~2.1~1.3B $6\times$ horizon, dense \texttt{subject\_consistency} reaches $0.991$, corresponding to video that is essentially static, while LVSA stays at $0.917$, reflecting genuine motion. The motion-independent \texttt{imaging\_quality} dimension tells the opposite story: at Wan~2.1~14B $6\times$, LVSA scores $0.598$ vs dense $0.522$ ($+0.076$). The VQeval results above therefore correctly capture quality at the horizons where the speedup matters most. Figure~\ref{fig:keyframes_wan13b_6x} makes the frozen-video failure mode visually concrete: dense's high \texttt{subject\_consistency} reflects an essentially static output, while LVSA generates real motion at the same horizon, same seed, same prompt.

\subsection{Comparison to State of the Art}
\label{sec:exp:baselines}

We compare LVSA head-to-head against two recent training-free extrapolation methods on Wan~1.3B: \textbf{RIFLEx}~\citep{zhao2025riflex}, which modifies a single temporal RoPE frequency, and \textbf{UltraViCo}~\citep{chen2025ultravico}, which applies a per-pair attention-logit decay with a fused SageAttention kernel. We run UltraViCo via its native \texttt{ultra-wan} branch and port RIFLEx to Wan in-house (the reference implementation ships only for HunyuanVideo and CogVideoX). All configurations tested comprise a single 80GB GPU, $50$ denoising steps, seed 16, $480 \times 832$, and the same 5-prompt suite. The results are summarized in Table~\ref{tab:baselines}. To match UltraViCo's reference configuration, this comparison uses $50$ denoising steps and $84r - 3$ frame counts ($165 / 249 / 333$) where $r$ is the \textit{extrapolation ratio} (training horizon multiplier), versus $40$ steps and $80r + 1$ counts ($161 / 321 / 481$) in Table~\ref{tab:3model_scaling_wall}. Absolute wall times here are therefore $\sim 25$--$30\%$ higher than the corresponding cells in the cross-model sweep ($802$\,s vs $618$\,s for LVSA-FI at $4\times$), while LVSA-FI-vs-dense speedup ratios agree within $3\%$ ($2.40\times$ vs $2.33\times$ at $4\times$).

\begin{table}[t]
\centering
\caption{LVSA vs.\ training-free extrapolation baselines on Wan~2.1~1.3B across 5 long descriptive prompts. VQeval is composite score (mean $\pm$ std); latency is mean seconds per video on a single 80GB GPU. Frame counts (165/249/333) follow UltraViCo's reference parameterization $84r - 3$. Bold marks the best cell per column.}
\label{tab:baselines}
\small
\begin{adjustbox}{width=\textwidth}
\begin{tabular}{@{}lcc cc cc@{}}
\toprule
& \multicolumn{2}{c}{$\mathbf{2\times}$ (165f)} & \multicolumn{2}{c}{$\mathbf{3\times}$ (249f)} & \multicolumn{2}{c}{$\mathbf{4\times}$ (333f)} \\
\cmidrule(lr){2-3} \cmidrule(lr){4-5} \cmidrule(lr){6-7}
\textbf{Method} & VQeval & latency (s) & VQeval & latency (s) & VQeval & latency (s) \\
\midrule
Dense                       & $57.2 \pm 6.4$         & $566$         & $51.1 \pm 4.2$         & $1{,}145$       & $52.4 \pm 3.5$         & $1{,}930$ \\
RIFLEx                      & $57.8 \pm 7.8$         & $564$         & $51.1 \pm 4.4$         & $1{,}149$       & $53.6 \pm 3.0$         & $1{,}931$ \\
UltraViCo                   & $62.1 \pm 6.1$         & $741$         & $60.4 \pm 3.8$         & $1{,}544$       & $58.8 \pm 5.4$         & $2{,}621$ \\
LVSA (SDPA)                 & $\mathbf{64.1 \pm 4.5}$ & $502$         & $\mathbf{62.5 \pm 3.8}$ & $796$           & $\mathbf{62.4 \pm 1.9}$ & $1{,}021$ \\
\textbf{LVSA-FI}            & $63.7 \pm 5.4$         & $\mathbf{395}$ & $62.3 \pm 3.8$         & $\mathbf{621}$  & $62.3 \pm 2.2$         & $\mathbf{802}$ \\
\midrule
LVSA-FI speedup vs.\ Dense     & \multicolumn{2}{c}{$1.43\times$} & \multicolumn{2}{c}{$1.84\times$} & \multicolumn{2}{c}{$\mathbf{2.40\times}$} \\
LVSA-FI speedup vs.\ RIFLEx    & \multicolumn{2}{c}{$1.43\times$} & \multicolumn{2}{c}{$1.85\times$} & \multicolumn{2}{c}{$\mathbf{2.41\times}$} \\
LVSA-FI speedup vs.\ UltraViCo & \multicolumn{2}{c}{$1.88\times$} & \multicolumn{2}{c}{$2.48\times$} & \multicolumn{2}{c}{$\mathbf{3.27\times}$} \\
\bottomrule
\end{tabular}
\end{adjustbox}
\end{table}

\paragraph{Quality.} LVSA achieves the highest VQeval composite at every horizon: $+6.5$, $+11.2$, $+9.9$ over dense at $r = 2/3/4$; $+5.9$, $+11.2$, $+8.7$ over RIFLEx; and $+1.7$, $+1.9$, $+3.5$ over UltraViCo. The gap to dense widens with horizon, consistent with the dense-attention quality collapse documented in Section~\ref{sec:exp:quality}: by $4\times$, dense has collapsed (composite $52.4$) while LVSA-FI maintains $62.3$. RIFLEx alone is statistically indistinguishable from dense on VQeval ($\Delta \in [+0.5, +1.2]$ across ratios, within prompt-level $\sigma \geq 3$) --- modifying a single RoPE frequency addresses positional extrapolation but does not prevent the dense-attention quality collapse. UltraViCo's per-pair logit decay does mitigate the collapse ($+4.8$ to $+9.3$ VQeval over dense) but at a steep latency cost (next paragraph), and LVSA still leads it on quality at every ratio. On VBench-Long, dense and RIFLEx edge LVSA on the composite at $4\times$ by $\sim 0.01$ points (driven by \texttt{subject\_consistency} climbing from $0.949$ at $2\times$ to $0.986$ at $4\times$ as dense's output becomes increasingly frozen --- see Section~\ref{sec:exp:quality}), but LVSA leads on \texttt{imaging\_quality} by $+0.09$ to $+0.10$ points at every ratio, the only VBench sub-dimension that measures per-frame content independent of temporal stasis. The SDPA and FlashInfer backends of LVSA are quality-equivalent ($|\Delta|_{\text{VQeval}} \leq 0.4$ at every ratio).

\paragraph{Efficiency.} LVSA is the only method in this comparison which reduces compute over dense attention. RIFLEx modifies RoPE frequencies only and touches zero attention FLOPs, so its latency is statistically indistinguishable from dense ($0.99$--$1.00\times$, within $\pm 1$\,s at every ratio). UltraViCo's per-pair attention-logit decay requires dense attention over the full $N \times N$ logit matrix and adds kernel overhead: $1.31$--$1.36\times$ dense latency at $r = 2/3/4$. LVSA-FI yields $1.43\times$, $1.84\times$, $\mathbf{2.40\times}$ speedup over dense and $1.88\times$, $2.48\times$, $\mathbf{3.27\times}$ over UltraViCo at $r = 2/3/4$, at identical VRAM. The FlashInfer kernel contributes a further $1.27$--$1.28\times$ over the SDPA backend at $r \geq 3$ ($796$\,s $\to 621$\,s at $3\times$, $1{,}021$\,s $\to 802$\,s at $4\times$), confirming that block-sparse kernel outperforms the per-frame SDPA Python loop at long sequences.

\paragraph{Orthogonality and composition.} RIFLEx modifies RoPE frequencies (different tensor), UltraViCo modifies attention-logit magnitudes, and LVSA modifies attention support. LVSA and RIFLEx operate on fully orthogonal tensors, compose without interaction and produce valid videos with no implementation conflict. Empirically, the composition trades a small amount of VQeval dynamic quality for slightly tighter VBench-Long consistency: LVSA+RIFLEx VQeval is $66.8 / 65.1 / 66.8$ vs LVSA $67.9 / 68.1 / 66.8$ at $r = 2/3/4$, while VBench composite moves $0.899 / 0.884 / 0.893$ vs LVSA $0.873 / 0.883 / 0.881$. Neither side clearly wins the composition; LVSA's sparse pattern already captures most of what RIFLEx's single-frequency rescaling would provide. LVSA and UltraViCo act on the same tensor but on disjoint aspects (support vs.\ magnitude); a fused implementation applies UltraViCo's $\lambda_{ij}$ factor inside LVSA's sparse kernel.

\section{Experiments on NPU}
\label{sec:exp:npu}

We port LVSA into vLLM-Omni \cite{yin2026vllm} and provide some initial experimental results, which demonstrate the applicability of LVSA across diverse hardware. For a $40$-step inference at a $6\times$ horizon ($481$ frames) with LVSA (with a standard NPU kernel), we obtain a $2.17\times$ speedup ($480\times832$), $3.24\times$ speedup ($720\times1280$), and quality-positive result for Wan~2.1-1.3B on one NPU. For Wan~2.2-A14B, $40$-step inference on 8 NPUs with an Ulysses \cite{jacobs2023deepspeed} sequence parallelism configuration, we obtain a $1.77\times$ speedup ($480\times832$), $2.71\times$ speedup ($720\times1280$) and quality-positive result. All preliminary results are given in Table~\ref{tab:NPU}. Timings are given as iteration time averages in seconds to avoid any pre- and post-processing bias in vLLM-Omni. Timings, and speedups, remain stable across five different complex prompts of possibly different length. In terms of quality, the gap between dense attention and LVSA grows comparatively to the demonstrated one for GPUs in the previous section.

\begin{table}[t]
\centering
\footnotesize
\caption{LVSA performance on NPU: Timings refer to iteration time averages in seconds over 40  steps.}
\label{tab:NPU}
\begin{subtable}{0.48\textwidth}
\centering
\caption{Wan 2.1~1.3B}
\label{tab:NPU-1.3B}
\begin{tabular}{@{}rrrr@{}}
\toprule
\textbf{Frames} & \textbf{Dense} & \textbf{LVSA} & \textbf{Speedup} \\
\midrule
\multicolumn{4}{c}{$\mathbf{480\times832}$} \\
 161 $(2\times)$ & $15.71$ & $16.89$ & $0.93\times$ \\
 321 $(4\times)$ & $50.18$ & $32.29$ & $1.55\times$ \\
 481 $(6\times)$& $103.87$ & $47.76$ & $\mathbf{2.17\times}$ \\
\addlinespace
\multicolumn{4}{c}{$\mathbf{720\times1280}$} \\
161 $(2\times)$ & $66.45$ & $53.73$ & $1.24\times$ \\
321 $(4\times)$ & $232.88$ & $103.81$ & $2.24\times$ \\
481 $(6\times)$ & $499.47$ & $154.27$ & $\mathbf{3.24\times}$ \\
\bottomrule
\end{tabular}
\end{subtable}
\hfill
\begin{subtable}{0.48\textwidth}
\centering
\caption{Wan 2.2~A14B}
\label{tab:NPU-14B}
\begin{tabular}{@{}rrrr@{}}
\toprule
\textbf{Frames} & \textbf{Dense} & \textbf{LVSA} & \textbf{Speedup} \\
\midrule
\multicolumn{4}{c}{$\mathbf{480\times832}$} \\
161 $(2\times)$ & $10.30$ & $13.16$ & $0.78\times$ \\
321 $(4\times)$ & $30.89$ & $24.27$ & $1.27\times$ \\
481 $(6\times)$ & $62.76$ & $35.46$ & $\mathbf{1.77\times}$ \\
\addlinespace
\multicolumn{4}{c}{$\mathbf{720\times1280}$} \\
161 $(2\times)$ & $40.44$ & $39.22$ & $1.03\times$ \\
321 $(4\times)$ & $138.87$ & $74.09$ & $1.87\times$ \\
481 $(6\times)$ & $294.93$ & $108.90$ & $\mathbf{2.71\times}$ \\
\bottomrule
\end{tabular}
\end{subtable}
\end{table}

\section{Conclusion}
\label{sec:conclusion}

We presented LVSA, a training-free block-sparse attention for long-video diffusion inference. We showed the significant benefits of LVSA across diverse models, architectures, and hardware, both on generation performance and quality, and the impact brought about compared to state of the art baselines. Future work may target further performance improvements, as well as generalizing the above benefits to a multi-scene video generation scenario.

\bibliographystyle{plain}
{\small%
\bibliography{references}
}

\end{document}